\documentclass[letterpaper, 10 pt, conference]{ieeeconf}  

\IEEEoverridecommandlockouts                              

\usepackage{etoolbox}
\makeatletter
\patchcmd{\@makecaption}
  {\scshape}
  {}
  {}
  {}
\makeatother

\usepackage{amsmath} 
\usepackage{amssymb}  
\usepackage{algorithmic}
\usepackage{siunitx}  
\usepackage{hyperref}
\usepackage{graphicx}

\newcommand{\vx}{\mathbf{x}}    
\newcommand{\vu}{\mathbf{u}}    

\newcommand{\seqX}{\mathbf{X}}    
\newcommand{\seqU}{\mathbf{U}}    
\usepackage{xcolor}

\newcommand{\RR}{\mathbb{R}}   

\newcommand{\sX}{\mathcal{X}}   
\newcommand{\sU}{\mathcal{U}}   
\newcommand{\sM}{\mathcal{M}}   

\newcommand{\vf}{\mathbf{f}}    

\DeclareMathOperator{\step}{step}

\newtheorem{definition}{Definition}

\usepackage[ruled,vlined,linesnumbered]{algorithm2e}
\SetInd{0.25em}{0.5em}
\SetAlFnt{\footnotesize\sf}
\SetKwRepeat{Do}{do}{while}
\SetKw{KwStep}{step}
\SetKwInput{KwData}{Input}
\SetKwInput{KwResult}{Result}
\SetKwBlock{RepeatForever}{repeat}{}
\SetKw{Continue}{continue}
\SetKwComment{Comment}{$\triangleright$\ }{}
\SetCommentSty{itshape}

\SetVlineSkip{0.2em}
\SetAlgoSkip{}

\SetAlCapHSkip{0cm}

\SetAlgoCaptionLayout{small}

\SetArgSty{textnormal}

\makeatletter
\patchcmd\algocf@Vline{\vrule}{\vrule \kern-0.4pt}{}{}
\patchcmd\algocf@Vsline{\vrule}{\vrule \kern-0.4pt}{}{}
\makeatother

\usepackage[capitalise]{cleveref} 

\DeclareRobustCommand{\abbrevcrefs}{%
\crefname{algorithm}{Alg.}{Algs.}%
}

\DeclareRobustCommand{\cshref}[1]{{\abbrevcrefs\cref{#1}}}

\title{\LARGE \bf
   iDb-RRT: Sampling-based Kinodynamic Motion Planning with Motion Primitives and Trajectory Optimization
}

\author{Joaquim Ortiz-Haro$^{1,2}$, 
Wolfgang Hönig$^{2}$, Valentin N. Hartmann$^{2,3}$,
Marc Toussaint$^{2}$,
Ludovic Righetti$^{1}$
\thanks{Website: \footnotesize\url{https://quimortiz.github.io/idbrrt/}}%
\thanks{
Code is available at Dynoplan (\footnotesize\url{https://github.com/quimortiz/dynoplan}) and 
Dynobench
(\footnotesize\url{https://github.com/quimortiz/dynobench}).}%
\thanks{$^{1}$Machines in Motion Laboratory, New York University, USA, $^{2}$TU Berlin, Germany, $^{3}$Computational Robotics Lab, ETH Zurich, CH.
This work was in part supported by the National Science Foundation grants 1932187, 2026479, 2222815 and 2315396.}
}

\newcommand{\ALGrrt}{\texttt{Geo-RRT-TO}\,}

\usepackage{amsmath}
\usepackage{amsfonts}
\usepackage{siunitx}
\usepackage{booktabs}

\begin{document}

\maketitle
\thispagestyle{empty}
\pagestyle{empty}

\begin{abstract}

	Rapidly-exploring Random Trees (RRT) and its variations have emerged as a robust and efficient tool for finding collision-free paths in robotic systems.
	However, adding dynamic constraints makes the motion planning problem significantly harder, as it requires solving two-value boundary problems (computationally expensive) or propagating random control inputs (uninformative).
	Alternatively, Iterative Discontinuity Bounded A* (iDb-A*), introduced in our previous study, combines search and optimization iteratively.
	The search step connects short trajectories (motion primitives) while allowing a bounded discontinuity between the motion primitives, which is later repaired in the trajectory optimization step.

	Building upon these foundations, in this paper, we present iDb-RRT, a sampling-based kinodynamic motion planning algorithm that combines motion primitives and trajectory optimization within the RRT framework.
	iDb-RRT is probabilistically complete and can be implemented in forward or bidirectional mode.
	We have tested our algorithm across a benchmark suite comprising 30 problems, spanning 8 different systems, and shown that iDb-RRT can find solutions up to 10x faster than previous methods, especially in complex scenarios that require long trajectories or involve navigating through narrow passages.

	%
	%
	%
	%
	%
	%

	%
	%

\end{abstract}

\section{Introduction}

Kinodynamic motion planning is a fundamental problem in robotics where the goal is to find collision-free trajectories in high-dimensional, continuous, and non-convex spaces, while also considering actuation limits and dynamics of the robot.
Over the last two decades, a wide variety of sampling-, search-, and optimization-based methods have been proposed to address (kinodynamic) motion planning problems.

A breakthrough was the introduction of Rapidly-exploring Random Trees (RRT) \cite{lavalle1998rapidly}, a sampling-based method that incrementally builds a tree of configurations by expanding nodes towards randomly sampled new configurations.
RRT-like algorithms (e.g., \cite{karaman2011sampling, kuffner2000rrt, bohlin2001randomized, gammel2014informed, RABITstar, gammell2015batch}) are highly efficient for geometric planning, i.e., motion planning settings that involve only joint configurations of the system, since in the geometric setting, two configurations can be connected exactly by using linear interpolation.

Although RRT-like algorithms can be adapted for kinodynamic motion planning (e.g., \cite{webb2012kinodynamic, long2020fuzzy}), their efficiency significantly decreases, as they typically require solving multiple two-point boundary value problems or the propagation of random control inputs.
Two-point boundary problems, as they arise for most robotic systems, often do not have an analytic solution, and solving them is computationally expensive, generally requiring the solution of a nonlinear trajectory optimization problem.
Propagating random control inputs tends to be uninformative for many systems, as random controls can lead to poor exploration of the state space, particularly in highly nonlinear systems such as quadrotors where random inputs often lead to instability in the system.
Further, it is not clear how to perform a bidirectional search, as in RRT-Connect  \cite{kuffner2000rrt}, in the kinodynamic setting with the propagation of random inputs.



\begin{figure}
	\centering
	\begin{tabular}{cc}
		\includegraphics[width=0.235\textwidth]{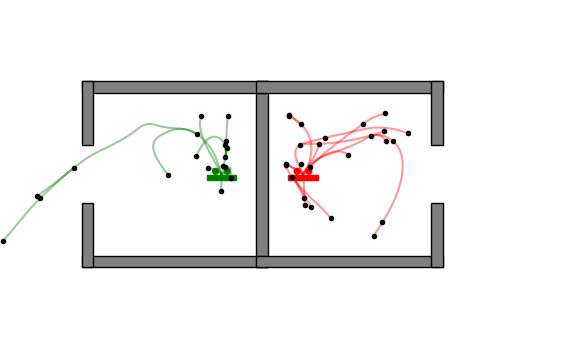} &
		\includegraphics[width=0.235\textwidth]{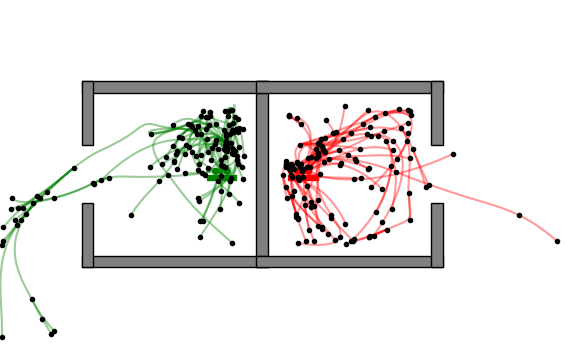}                         \\
		{\footnotesize (a)}                                                        & {\footnotesize (b)} \\
		\includegraphics[width=0.235\textwidth]{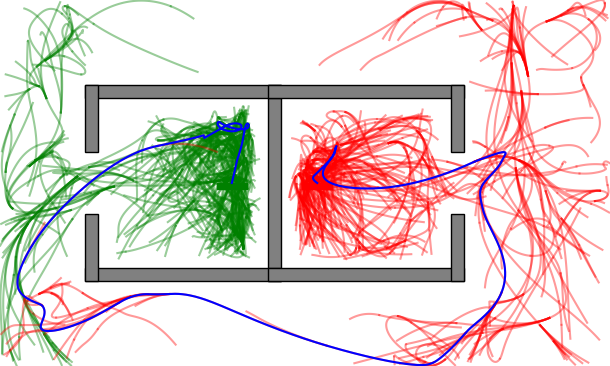} &
		\includegraphics[width=0.235\textwidth]{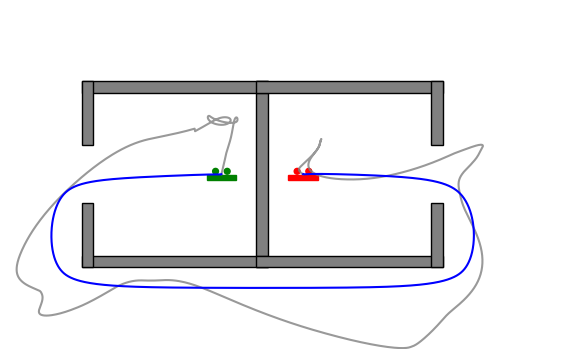}                    \\
		{\footnotesize (c)}                                                        & {\footnotesize (d)}
	\end{tabular}
	\caption{iDb-RRT combines a forward or bidirectional RRT search with motion primitives (Db-RRT) and trajectory optimization iteratively. (a,b) In the search step, the RRT is expanded by connecting motion primitives with a bounded discontinuity. (c) The output of the RRT is a trajectory with a bounded discontinuity in the dynamics constraints. (d) Using trajectory optimization, we generate a dynamically feasible trajectory.
		Problem visualization: \textit{Planar Rotor in Double bugtrap}.
		\vspace{-.3cm}
	}
\end{figure}

Alternative approaches for kinodynamic motion planning are optimization-based methods \cite{TrajOpt, GuSTO, malyutaConvexOptimizationTrajectory2021}, which scale polynomially instead of exponentially but require an initial guess and may fail to converge; and search-based methods \cite{PivtoraikoThesis, pivtoraikoKinodynamicMotionPlanning2011}, which provide strong theoretical guarantees but require a pre-defined discretization of the state or control space.
More recently, hybrid methods have been proposed to merge the strengths of the three previous approaches to kinodynamic motion planning \cite{natarajanInterleavingGraphSearch2021, sakcakSamplingbasedOptimalKinodynamic2019, DIRT, kamat2022bitkomo, natarajan2024pinsat}.
Iterative Discontinuity-Bounded A* (iDb-A*) \cite{ortizharo2023idba} introduces an approach based on A*-search with \emph{motion primitives}, i.e., short and locally optimal trajectories, that are connected not necessarily exactly, but allowing for a bounded discontinuity between primitives.
These discontinuities between the motion primitives are later rectified using trajectory optimization (TO).
By iteratively combining optimization and search with an increasing number of motion primitives and a reduced discontinuity bound, this method achieves asymptotically optimal motion planning and outperforms state-of-the-art methods across various robotic systems.
A primary limitation of iDb-A* is its inefficiency in finding an initial solution, particularly in large environments, where the time required to find the initial solution remains high.

In this paper, we combine the strengths of the exploration of RRT with the concept of discontinuities between motion primitives and trajectory optimization.
We present iDb-RRT (\underline{i}terative \underline{D}iscontinuity-\underline{b}ounded RRT), a new kinodynamic motion planning algorithm that builds on the ideas of allowing discontinuities in an initial motion from iDb-A*, and integrates the RRT exploration strategy with short motion primitives and trajectory optimization.
iDb-RRT samples a random configuration, then expands the configuration that is closest using applicable motion primitives with bounded discontinuity.
Once a solution is found, we employ trajectory optimization to correct the discontinuities between motion primitives.
By incrementally increasing the number of primitives and reducing the allowed discontinuity, our algorithm achieves probabilistic completeness.

We analyze both a forward and a bidirectional version of iDb-RRT.
In the open-source benchmark \textit{Dynobench}  comprising 30 problems across 8 different systems, iDb-RRT significantly outperforms state-of-the-art methods in initial solution time, especially in complex scenarios requiring long-horizon planning or navigating through narrow passages.

\section{Related Work}

In this section, we discuss previous work on RRTs for kinodynamic motion planning and methods combining sampling and optimization.
A more comprehensive review of methods in kinodynamic motion planning can be found in \cite{ortizharo2023idba, masoud2010kinodynamic}.

Sampling-based methods often grow the search tree towards a randomly sampled configuration by solving two-point boundary value problems \cite{kinodynamicRRT} to connect two states precisely, or by propagating random control inputs \cite{shomeAsymptoticallyOptimalKinodynamic2021b}.
Previous work has focused on improving the expansion step (also called steering function) for specific systems \cite{webb2012kinodynamic, LQR-RRTstar, goretkin2013optimal}, better exploration by most informative sampling \cite{SSTstar, tang2020vector}, better heuristics \cite{wang2022efficient}, better integration of nonlinear solvers as a subroutine in sampling-based planners \cite{RABITstar, stoneman2014embedding}, or using motion primitives \cite{sakcakSamplingbasedOptimalKinodynamic2019} in a discretized configuration space.
Compared to the previously discussed methods, our approach plans with the full dynamics (with bounded discontinuity), does not require discretization of the workspace, and does not require solving two-point boundary value problems in the RRT expansion step.
This is enabled by leveraging precomputed motion primitives and allowing discontinuities in the planning stage, which are later fixed using trajectory optimization (TO).

Leveraging TO is a common approach for both geometric \cite{kamat2022bitkomo, RABITstar} and kinodynamic motion planning, e.g., as a final post-processing step to improve cost and smoothness \cite{ravankar2018path}.

In kinodynamic planning, previous work often involves planning using a simplified geometric model \cite{bortoff2000path, allen2019real, leu2022long} and tracking the resulting reference using trajectory optimization or an optimization-based controller.
This approach is commonly used in high-dimensional systems, e.g., \cite{bry2015aggressive, wahba2023kinodynamic} for UAVs or \cite{jelavic2023lstp, bellicoso2018dynamic} for legged robots.
Unfortunately, initially using a simplified model and accounting for the full dynamics later is limiting and might lead to infeasible optimization problems if the initial guess is not close to a dynamically feasible trajectory \cite{li2021model}.

We also use TO for computing the final feasible trajectory, but we plan with the full dynamics (with bounded discontinuity).
As this discontinuity can be made arbitrarily low, and optimization and search are combined iteratively, iDb-RRT is probabilistically complete under mild assumptions.

\section{Problem Definition}

We consider a robot with a continuous state \(\vx \in \sX\) (e.g., \(\sX \subseteq \mathbb{R}^{d_x}\)) and a control vector \(\vu \in \sU \subset \mathbb{R}^{d_u}\).
The dynamics of the robot are deterministic, described by a differential equation, \begin{equation} \dot{\vx} = \vf(\vx, \vu).
	\label{eq:dynamics}
\end{equation}
To employ gradient-based optimization, we assume that we can compute the Jacobian of \(\vf\) with respect to \(\vx\) and \(\vu\), typically available in systems studied in kinodynamic motion planning, such as mobile robots or rigid-body articulated systems.
We use \(\mathcal{X}_{\text{free}} \subseteq \mathcal{X}\) to denote the collision-free space, i.e., the subset of states that are not in collision with the obstacles in the environment.

We discretize the dynamics \eqref{eq:dynamics} with a zero-order hold, i.e., we assume the applied control is constant during a time step of duration \(\Delta t\).
The discretized dynamics can then be written as, \begin{equation} \label{eq:dynamics_discrete} \vx_{k+1} \approx \text{step}(\vx_k, \vu_k) \equiv \vx_k + \vf(\vx_k, \vu_k)\Delta t \,, \end{equation} using a small \(\Delta t\) to ensure the accuracy of the Euler approximation.
We use \(K \in \mathbb{N}\) to denote the number of time steps (which is not fixed but subject to optimization), \(\seqX = \langle \vx_0, \vx_1, \ldots, \vx_K \rangle\) to denote the sequence of states sampled at times \(0, \Delta t, \dots, K\Delta t\) and \(\seqU = \langle \vu_0, \vu_1, \ldots, \vu_{K-1} \rangle\) to denote the sequence of controls applied to the system for the time frames \([0,\Delta t), [\Delta t, 2\Delta t), \ldots, [(K-1)\Delta t, K\Delta t)\).
The objective of navigating the robot from its start state \(\vx_s\) to a goal state \(\vx_g\) can then be framed as the search problem, \begin{subequations} \label{eq:motion-planning} \begin{align}                               & \text{find } {\seqU,\seqX,K} \label{eq:j}                                                                      \\ \text{s.t.
                                          } & \vx_{k+1} = \text{step}(\vx_k, \vu_k)                     & \forall k \in \{0,\ldots,K-1\} \label{eq:step} \,, \\
                                            & \vu_k \in \sU                                             & \forall k \in \{0,\ldots,K-1\} \label{eq:u} \,,    \\
                                            & \vx_k \in \sX_{\text{free}} \subseteq \sX                 & \forall k \in \{0,\ldots,K\} \label{eq:x} \,,      \\
                                            & \vx_0 = \vx_s; \,\, \vx_K = \vx_g \label{eq:terminal} \,.
	\end{align}
\end{subequations}

In this paper we focus on finding a valid trajectory quickly (i.e, very little compute time), as opposed to finding the optimal solution.
Although there is no explicit minimization of a cost function in our algorithms, we can evaluate the cost of the trajectory a posteriori.
We use the cost term \(J(\mathbf{U},\mathbf{X}) = \sum_{k=0}^{K-1} j(\vu_k,\vx_k)\, \Delta t\), with \(j(\vu_k,\vx_k) = 1\) for minimal time (span) (alternatively, one might use \(j(\vu_k,\vx_k) = \|\vu_k\|^2\) for minimal control effort).
%
We assume the dynamics function \(\text{step}(\vx,\vu)\), control space \(\mathcal{U}\), state space \(\mathcal{X}\), and cost function \(j(\vx,\vu)\), are known before solving the problem, which allows us to precompute motion primitives.

\section{iDb-RRT}

\subsection{Background}

Our approach relies on two concepts, that we now define.

\begin{definition}[Discontinuity Bounded Solution]
	\label{def:discontinuity-bounded-solution}
	A trajectory \( \seqX = (\vx_0, \ldots ,\vx_K) , \seqU = (\vu_0, \ldots, \vu_{K-1}) \) is a $\delta$-discontinuity bounded solution of the kinodynamic motion planning problem \cref{eq:motion-planning} if: $d(\vx_{k+1}, \text{step}(\vx_k, \vu_k)) \leq \delta$, $d(\vx_{0}, \vx_s) \leq \delta$, $d(\vx_{K}, \vx_g) \leq \delta$, $\vx_k \in \sX_{\text{free}}$ and $\vu_k \in \sU$, where $d(\cdot,\cdot)$ is a distance function, e.g., a weighted Euclidean norm, and $\delta \ge 0$.
\end{definition}

The search step of our algorithm, Db-RRT, generates solutions that are discontinuity bounded, while the trajectory optimization step rectifies these solutions to satisfy \cref{eq:motion-planning}.

\begin{definition}[Motion Primitive]
	A motion primitive \( m = (\seqX, \seqU, \vx_s, \vx_f, c) \) is a sequence of states \( \seqX = (\vx_0, \ldots ,\vx_N) \), \( \vx_k \in \sX \), and controls \( \seqU = (\vu_0, \ldots, \vu_{N-1}) \), \( \vu_k \in \sU \) that fulfill the dynamics \( \vx_{k+1} = \step(\vx_k,\vu_k) \).
	It connects the start state \( \vx_s = \vx_0 \) and the final state \( \vx_f = \vx_N \), with a corresponding cost \( c \in \RR^+ \).
	The length of the motion primitive (i.e., the number of states and controls) is randomized.
	\label{def:motion-primitive}
\end{definition}

%
%

\begin{figure}
	\centering
	\includegraphics[width=.4\textwidth]{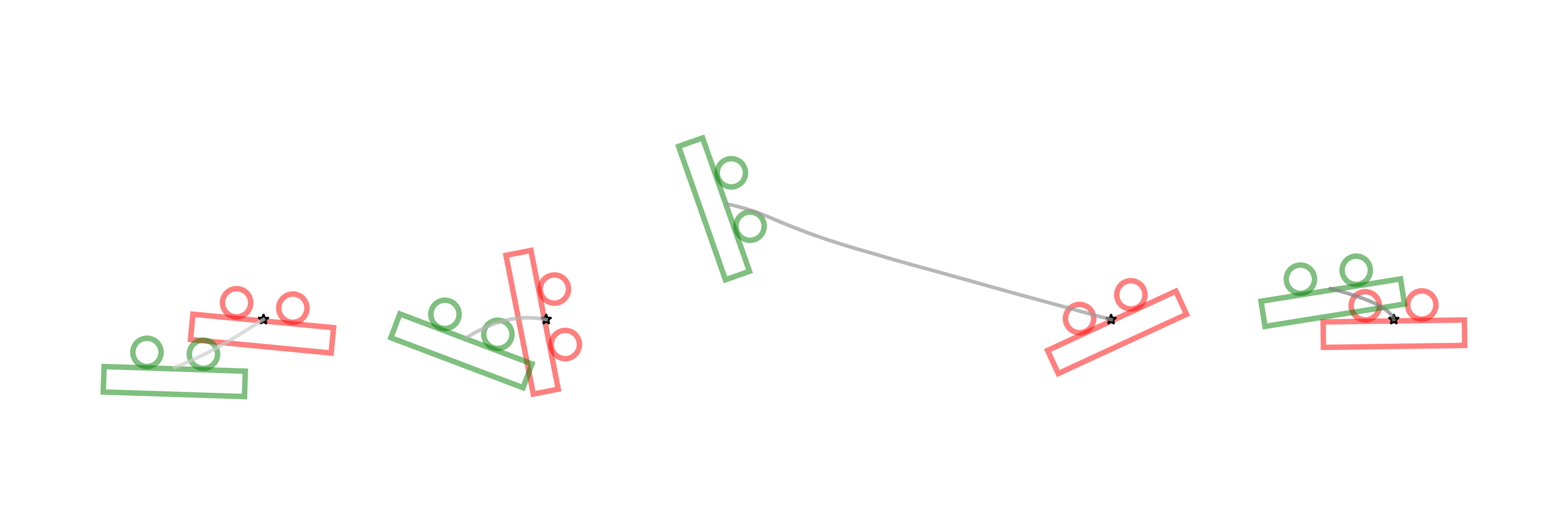}
	\includegraphics[width=.3\textwidth]{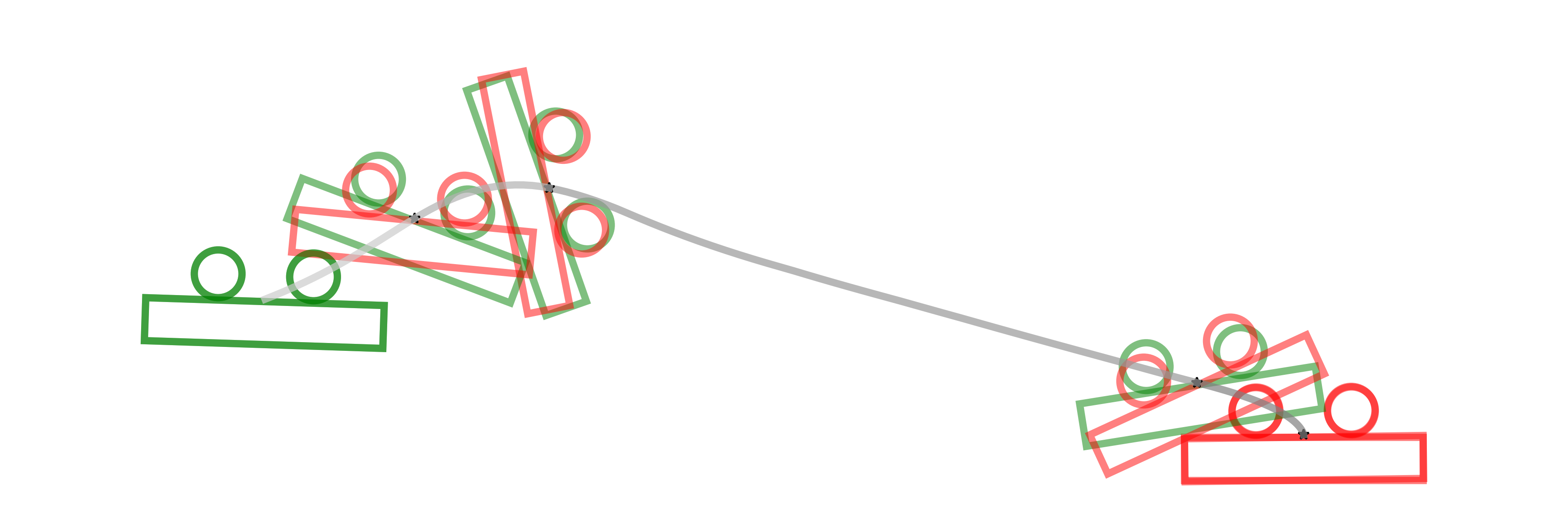}
	\caption{\textit{Top}: Four motion primitives in the system \textit{Planar rotor}.
		The initial state (green), final state (red) and duration are randomized.
		\textit{Bottom}: During the search step (Db-RRT), motion primitives are connected allowing for a bounded discontinuity.
		In this visualization, we connect these four motion primitives from left to right.
		The green and red configurations indicate the first and last configurations of each primitive.
		Note that their rotation component does not match exactly (further, discontinuities in the velocity components are not shown).
	}
	\label{fig:primitives}
\end{figure}

A large set of motion primitives can be generated offline by sampling random start and goal states, and attempting to connect them using nonlinear trajectory optimization algorithms.
This results in a superior distribution of primitives in terms of coverage of the state space, compared to propagating random control inputs, and it guarantees asymptotic coverage of the state space \cite{ortizharo2023idba}.
Importantly, we can later use known properties of the system to adapt primitives on-the-fly to match a state during the search, e.g., by using translation invariance of mobile robots, we can \emph{translate} a primitive to match the position components of the state space \cite{ortizharo2023idba}.
\cref{fig:primitives} displays four motion primitives in the system \textit{planar rotor} and how they can be connected with a bounded discontinuity.

\subsection{Overview}

Our approach is summarized in \cshref{alg:overview}.
We assume that a large set of motion primitives \( \sM_{\mathrm{L}} \) has been precomputed and is available before planning.
iDb-RRT iteratively runs two steps until the first valid solution is found:

\begin{itemize}
	\item An RRT search algorithm that connects motion primitives with bounded discontinuity, called Db-RRT.
	      The output is a discontinuity bounded solution, i.e., a collision-free trajectory with bounded violation of dynamic constraints (\cref{def:discontinuity-bounded-solution}).

	\item Gradient-based trajectory optimization, which attempts to repair the discontinuities between the motion primitives to produce a dynamically feasible trajectory.
\end{itemize}

If the search fails to find a solution within a given timeout (TerminateCondition), we increase the number of available motion primitives.
If gradient-based optimization fails, we reduce the allowed discontinuity.
In practice, we typically require only one or two outer iterations (that is, a call to the search and optimization algorithms) to find a solution.
We decrease the allowed discontinuity following a geometric sequence, \(d_i = d_{i-1} \cdot d_r\) with a fixed rate \(d_r<1\), and increase the number of primitives also following a geometric sequence \(m_i = m_{i-1} \cdot m_r\) with a fixed rate \(m_r>1\).


\begin{algorithm}[t]
	\caption{iDb-RRT -- Iterative Discontinuity Bounded RRT}
	\KwData{$\vx_s, \vx_g, \mathrm{step} , \sX_{\mathrm{free}}, \sU, \sM_{\mathrm{L}} $}
	\KwResult{$\seqX, \seqU$ }
	\label{alg:overview}
	\DontPrintSemicolon
	\SetKwFunction{AddPrimitives}{AddPrimitives}
	\SetKwFunction{ExtractPrimitives}{ExtractPrimitives}
	\SetKwFunction{ComputeDelta}{ComputeDelta}
	\SetKwFunction{ChooseDelta}{ChooseDelta}
	\SetKwFunction{ChoosePrimitives}{ChoosePrimitives}
	\SetKwFunction{DecreaseDelta}{DecreaseDelta}
	\SetKwFunction{IncreasePrimitives}{IncreasePrimitives}
	\SetKwFunction{dbrrt}{db-RRT}
	\SetKwFunction{Optimization}{Optimization}
	\SetKwFunction{Return}{Return}
	$\delta \leftarrow \delta_0$ \Comment*{Choose initial discontinuity bound}
	$\sM \leftarrow \ChoosePrimitives(\sM_{\mathrm{L}})$

	\Comment*{Choose initial subset of primitives from $\sM_{\mathrm{L}}$}
	\While{not found}{
		$\seqX_d, \seqU_d  \leftarrow$ \dbrrt{$\vx_s, \vx_g, \sX_{\mathrm{free}}, \sM, \delta$}\label{alg:overview:dbAstar}\;
		\If{$\seqX_d, \seqU_d$ successfully computed}{
			$\seqX, \seqU \leftarrow$ \Optimization{$\seqX_d, \seqU_d, \vx_s, \vx_g , \mathrm{step} , \sX_{\mathrm{free}} , \sU $}\label{alg:overview:opt}\;
			\If{$\seqX, \seqU$ successfully computed}{
				\Return{$\seqX, \seqU$} \Comment*{New solution found}
			}
			\Else{
				$\delta \leftarrow \DecreaseDelta(\delta)$
			}
		}
		\Else {
			$\delta \leftarrow \DecreaseDelta(\delta)$ \label{alg:overview:else1}\;
			$\sM \leftarrow \IncreasePrimitives(\sM, \sM_{\mathrm{L}})$ \label{alg:overview:else2}
		}
	}
\end{algorithm}

\subsection{Db-RRT: RRT with Motion Primitives}

Db-RRT is an RRT algorithm that connects motion primitives with bounded discontinuity, following the general RRT algorithm to choose the next state to expand.
This approach provides a Voronoi bias (i.e., nodes at the frontier of the search tree are more likely to get expanded), thus rapidly exploring the feasible state space.
In \cshref{alg:rrt}, we describe our Db-RRT algorithm and highlight our modifications from RRT.
In Db-RRT, the expansion operation is performed using motion primitives with bounded discontinuity.
Given the state \(\mathbf{x}_{\text{near}}\), we assess which primitives are applicable
(e.g., \cshref{alg:db-rrt-focused:expand} in 
\cshref{alg:rrt-togoal}). We then differentiate between focused expansion (\cshref{alg:rrt-togoal}), where we select the primitive that brings us closest to \(\mathbf{x}_{\text{rand}}\) from a finite number of nearby candidates, and uninformed expansion (\cshref{alg:rrt-torand}), where we choose one collision-free primitive at random.
With a small probability (the so-called goal bias), we expand towards the goal state instead of a random state.

We stop when we find a state that is within a distance lower than \(\delta\) of the goal state.
Further, the value of \(\delta\) is also used to avoid creating nodes in the tree that are too close to previously discovered nodes.

Both expansion strategies are guaranteed to find a solution, if one exists, given sufficient compute time.
The inherent trade-off is that \cshref{alg:rrt-togoal} requires more compute time, as it involves evaluating collisions for multiple motion primitives, but it provides a more focused and uniform expansion.
In our implementation, we utilize \cshref{alg:rrt-togoal} for expansions towards the goal and \cshref{alg:rrt-torand} for expansions towards random nodes, but any combination of these two approaches is valid.
Focused and uninformed expansion are analogous to guided Monte-Carlo and Monte-Carlo propagation in classic RRT literature, but in Db-RRT we use motion primitives instead of randomly sampled controls.

\begin{algorithm}[t]
	\caption{Db-RRT -- Rapidly-Exploring Random Trees with Motion Primitives}
	\KwData{$\vx_s, \vx_g, \sX_{\mathrm{free}}, \sM, \delta $}
	\KwResult{$\seqX_d, \seqU_d$ }
	\label{alg:rrt}
	\DontPrintSemicolon
	\SetKwFunction{AddPrimitives}{AddPrimitives}
	\SetKwFunction{ExtractPrimitives}{ExtractPrimitives}
	\SetKwFunction{ComputeDelta}{ComputeDelta}
	\SetKwFunction{ChooseDelta}{DecreaseDelta}
	\SetKwFunction{ChoosePrimitives}{IncreasePrimitives}
	\SetKwFunction{dbrrt}{db-RRT}
	\SetKwFunction{Optimization}{Optimization}
	\SetKwFunction{Return}{Return}
	\SetKwFunction{Sample}{Sample}
	\SetKwFunction{Nearest}{Nearest}
	\SetKwFunction{AddNode}{AddNode}
	\SetKwFunction{ExpandDb}{ExpandDb}
	\SetKwFunction{NearestDistance}{NearestDistance}
	\SetKwFunction{TracebackTrajectory}{TracebackTrajectory}
	\SetKwFunction{TerminateCondition}{TerminateCondition}

	$\mathcal{T} \leftarrow \AddNode(\vx_s) $\;
	\While{$\neg \TerminateCondition()$ }{

	\If{rand() $<$ goalBias}{
		$\vx_{\text{rand}} \leftarrow \vx_g$\;
	}
	\Else {
		$\vx_{\text{rand}} \leftarrow \Sample(\sX_\textup{free})$\;
	}

	$\vx_{\text{nearest}} \leftarrow \Nearest(\mathcal{T}, \vx_{\text{rand}})$\;
	{ \color{blue}
	$\vx_{\text{new}}, m \leftarrow \ExpandDb(\vx_{\text{nearest}}, \vx_{\text{rand}}, \sX_{\mathrm{free}}, \sM, \delta)$ \label{alg:rrt:change1}\;
	}
	\If{$\vx_{\text{new}} \neq \text{NULL}$}{

		\If{d($\vx_{\textup{new}}, \vx_g$) $<$  $\delta$}{
			$\mathcal{T} \leftarrow \AddNode(\mathcal{T}, \vx_{\text{new}} )$
   
		$\seqX_d, \seqU_d \leftarrow \texttt{TracebackTrajectory}(\mathcal{T}, \vx_{\text{new}})$\;
  
			\Return{$\seqX_d, \seqU_d$} \Comment*{Discontinuity bounded  solution}
		}

		\ElseIf  {  \color{blue} \NearestDistance($\mathcal{T}, \vx_{\textup{new}}$) $>$  $\delta$ \label{alg:rrt:change2}}{
			$\mathcal{T} \leftarrow \AddNode(\mathcal{T}, \vx_{\text{new}} )$
		}

	}
	}
\end{algorithm}


\subsection{Db-RRT-Connect and other Db-RRT variants}
The expansion step of Db-RRT (\cshref{alg:rrt-togoal,alg:rrt-torand}) can be integrated with many of the variations and enhancements of RRT that have been previously proposed,   

\paragraph{Backward and Bidirectional Search} Inspired by RRT-Connect \cite{kuffner2000rrt}, we present a bidirectional variant of Db-RRT,  where we grow two trees, one from the start (using standard motion primitives) and one from the goal (using reversed motion primitives), and attempt to connect them.
The expansion step in a backward search mirrors that of a forward search but requires reversing the order of states and controls in the motion primitives beforehand.
The two trees are connected if two of their states are within the discontinuity bound.

\paragraph{Asymptotically Optimal Algorithms}
Db-RRT can also be applied to RRT variants that require connecting two states precisely, instead of only expanding the state towards random targets.
The discontinuity bound \(\delta\) can be leveraged to consider two states as equivalent—thereby enabling their exact connection in any rewiring step, such as in RRT* \cite{karaman2011sampling}.
Such rewiring steps, which are essential for the asymptotic optimality of RRT* and its variants, are already implemented in iDb-A* \cite{ortizharo2023idba}.

\begin{algorithm}[t]
	\caption{Expand-Db: Focused}
	\label{alg:rrt-togoal}
	\SetKwFunction{Return}{Return}

	\KwData{$\vx_o, \vx_t,  \sX_{\mathrm{free}}, \sM, \delta$}
	\KwResult{$\vx, m$}

	$\sM_c \gets \texttt{NearestR}(\vx_o, \sM, \delta)$ \label{alg:db-rrt-focused:expand}\;

	$m_{\text{b}}  = NULL , \quad d_{\text{b}} = \infty$

	\For{$m \in \sM_c$}{
		\If{ $m \in \sX_{\text{free}} ~ \text{and} ~ d(m.\vx_f,\vx_t) < d_{\text{b}}$}{
			$m_{\text{b}} \gets m, \quad d_{\text{b}} \gets d(m.\vx_f,\vx_t)$

		}
	}
	\If{ $m_{\text{b}} \neq NULL$}{
	\Return $m_{b}.\vx_f,m_{b}$
	}
	\Return $NULL, NULL$
\end{algorithm}

\begin{algorithm}[t]
	\caption{Expand-Db: Randomized}
	\label{alg:rrt-torand}
	\SetKwFunction{Return}{Return}
	\SetKwFunction{RandomPermutation}{RandomPermutation}
	\KwData{$\vx_o, \vx_t,  \sX_{\mathrm{free}}, \sM, \delta$}
	\KwResult{$\vx, m$ }

	$\sM_c \gets \texttt{NearestR}(\vx_o, \sM, \delta)$


	\For{$m \in \text{RandomPermutation}(\sM_c)$}{
		\If{ $m \in \sX_{\text{free}}$}{
			\Return $m.\vx_f, m$
		}
	}
	\Return $NULL, NULL$

\end{algorithm}

%
%
%


\subsection{Trajectory Optimization}



%

The output of Db-RRT is a sequence of states and controls that connects the start and goal states with a bounded discontinuity, see \cref{def:discontinuity-bounded-solution}.
In the optimization step of iDb-RRT, we employ nonlinear trajectory optimization to repair the discontinuity between the motion primitives and to obtain a feasible and locally optimal trajectory.

For gradient-based trajectory optimization, we require the gradients of the dynamics and the cost function with respect to the states and controls.
These can be easily obtained for most robotics systems using finite differences, analytic expressions, or a differentiable simulator.
Instead of the binary collision check in Db-RRT, we now use a signed distance function.

In the trajectory optimization step, the number of time steps \(K\) is fixed by the output of Db-RRT.
If desired, we can also optimize the duration of the trajectory by including the length of the time interval in the optimization problem or using other techniques, as explored in \cite{ortizharo2023idba}.
Because our goal is to find a valid trajectory quickly, we choose not to include the time interval as an optimization variable.
This choice is supported by the fact that trajectories from RRT-like algorithms tend to be suboptimal, where the time duration of the initial guess is often sufficient to reach the goal.

To solve the trajectory optimization problem, we use the Differential Dynamic Programming (DDP) algorithm, which is a second-order method for solving optimal control problems of the form Eq. \eqref{eq:ddp}.
Collision and goal constraints, and state and control bounds of the original kinodynamic motion planning problem are added to the cost \eqref{eq:trajectory-optimization} with a squared penalty method and a max activation function for inequalities.
Further, we include small regularization terms on the control effort and the acceleration of the system to improve convergence.
\begin{subequations} \begin{align} \min_{ \seqX, \seqU } & \sum_{k=0}^{K-1} c(\vx_k,\vu_k) + c_K(\vx_K)\,, \label{eq:trajectory-optimization} \\ \text{s.t.
              } \quad               & \vx_{k+1} = \text{step}(\vx_{k},\vu_k) \quad \forall k \in \{0,\ldots,K-1\}\,,     \\
                                    & \vx_0 = \vx_s\,.
	\end{align}
 \label{eq:ddp}
\end{subequations}
In particular, we use the optimization algorithm \emph{Feasibility-driven DDP}~\cite{Crocoddyl}, which can be warm-started with an infeasible sequence of states and controls, providing a good balance between local convergence and globalization.




\subsection{Analysis}



The RRT algorithm is probabilistically complete \cite{kinodynamicRRT,kleinbortProbabilisticCompletenessRRT2019a}, that is, the probability of eventually finding a solution, if one exists, converges to one.
The proof assumes that the planning problem is $\delta_1$-robust (informally: the solution should not require traversing a "gap" smaller than $\delta_1$) and that the dynamics are Lipschitz continuous.
Formally, it uses an inductive argument over overlapping balls that cover the solution trajectory, demonstrating that the probability of finding an edge between neighboring balls is non-zero.

We first consider Db-RRT (\cref{alg:rrt}) with the precondition that we have a sufficiently large set of motion primitives $\sM_{\mathrm{L}}$ and a discontinuity bound $\delta < \delta_1$.
Then, the additional if-condition in \cref{alg:rrt:change2} does not prevent finding a solution.
\Cref{alg:rrt:change1} changes the distribution for the expansion operation but continues to assign a positive probability density to all successors for large sets of randomly generated $\sM_{\mathrm{L}}$.
Next, we consider iDb-RRT (\cref{alg:overview}).
If Db-RRT fails to find a solution because at least one precondition is violated (a large $\sM_{\mathrm{L}}$ and $\delta < \delta_1$), we adjust both parameters and repeat (\crefrange{alg:overview:else1}{alg:overview:else2}), yielding a non-zero probability of executing Db-RRT with parameters that fulfill our assumptions.
Finally, we assume that there exists a \(\delta\) such that if Db-RRT generates a \(\delta\)-discontinuity bounded solution, the trajectory optimization algorithm will converge with a non-zero probability, which makes our algorithm, iDb-RRT, probabilistically complete.



In practice, we demonstrate that we can use a large discontinuity $\delta$ and a small number of primitives to efficiently find solutions to a wide range of problems.

\section{Experiments}

We evaluate iDb-RRT on 30 problems that include 8 different dynamical systems in various environments.
The first 16 problems are inspired by previous work on kinodynamic motion planning \cite{SSTstar, shomeAsymptoticallyOptimalKinodynamic2021b, hoenig2022benchmarking, granados2022towards}
(\textit{selected problems} in \cite{ortizharo2023idba}, first \num{16} rows in \cref{tab:all}). Furthermore, we include 14 additional problems with the same dynamical systems but in larger, more complex environments with more obstacles, which require longer trajectories (last \num{14} rows in \cref{tab:all}).

All benchmark problems are available in \textit{Dynobench}.
It provides a C++ implementation of the dynamical systems (dynamics with analytical Jacobians, state, and bound constraints), collision and signed distance function (based on the Flexible Collision Library, FCL), the environments (in human-friendly YAML files), and visualization tools.

Implementations of iDb-RRT and the other planners are available in \textit{Dynoplan}, including the motion primitives and instructions to replicate the benchmark results.
Visualizations of the problems and examples of solution trajectories computed by our algorithm are available on our website.


\subsection{Dynamical Systems}



\begin{figure*}
	\label{fig:newproblems}
	\centering



	\begin{tabular}{ccccc}
		\includegraphics[width=0.12\textwidth]{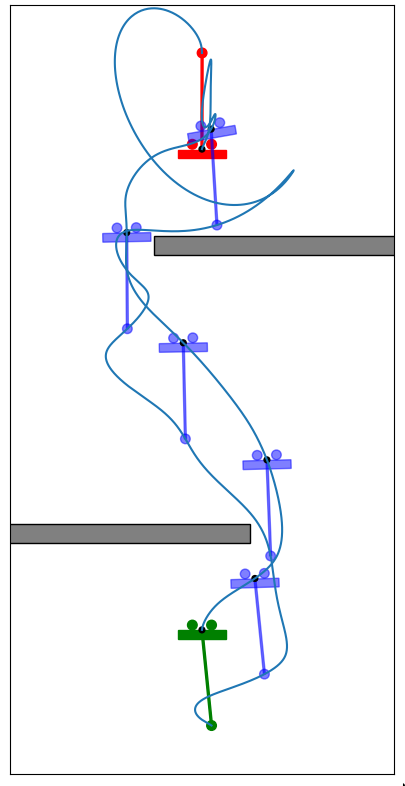} &
		\includegraphics[width=0.18\textwidth]{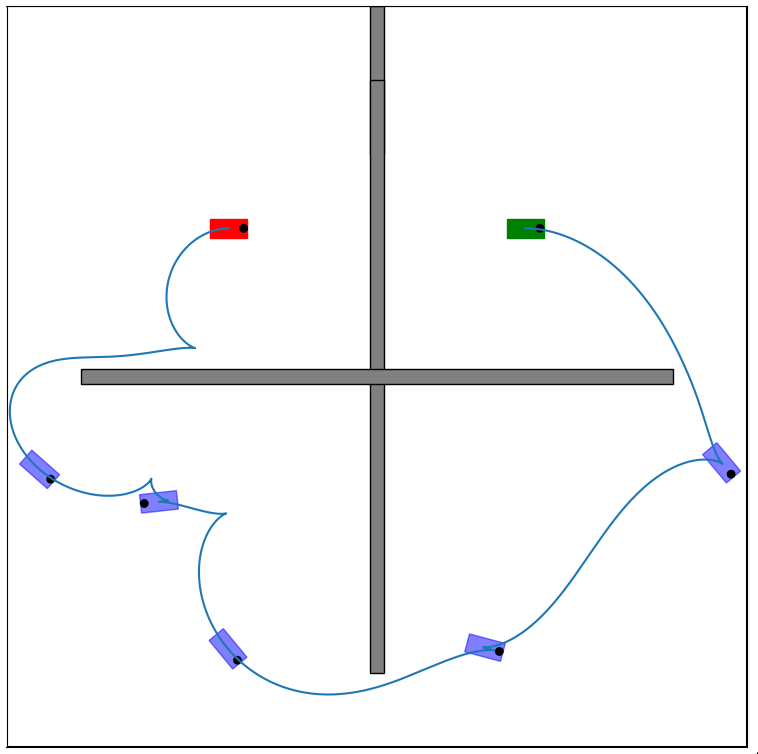} &
		\includegraphics[width=0.18\textwidth]{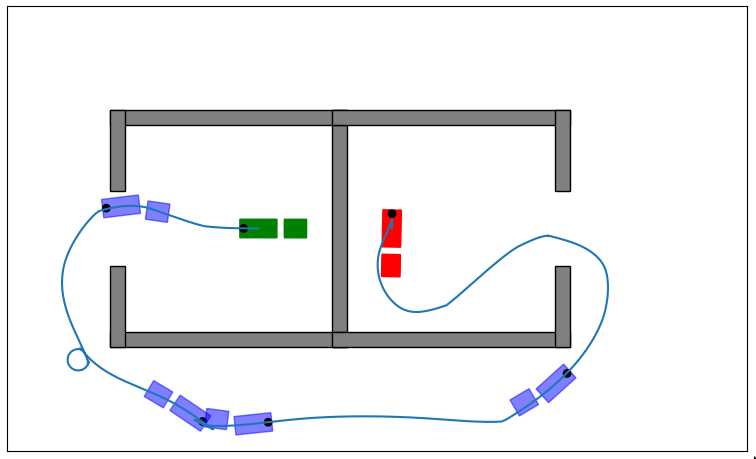} &
		\includegraphics[width=0.18\textwidth]{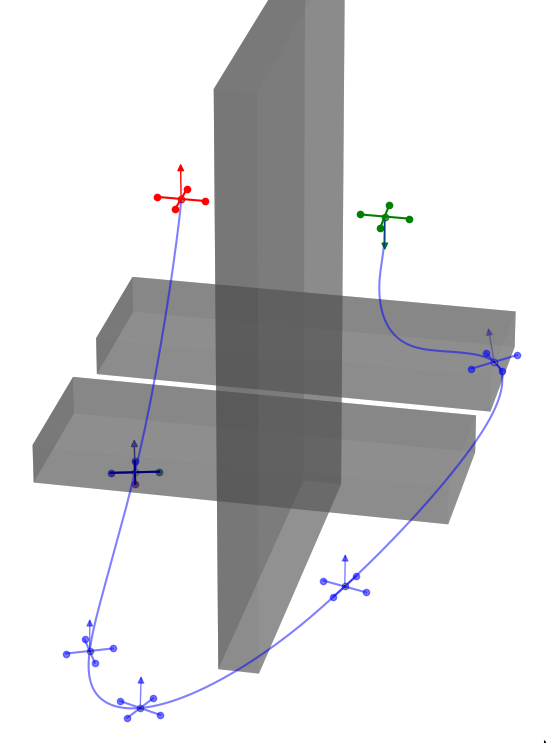} &
		\includegraphics[width=0.18\textwidth]{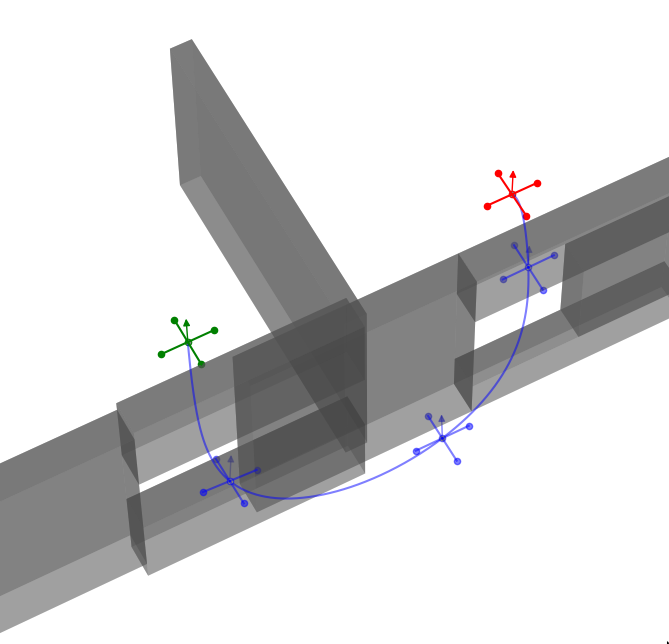}                                                 \\

		{\footnotesize (a)}                                                                & { \footnotesize (b)} & { \footnotesize (c)} &
		{ \footnotesize (d)}                                                               &
		{ \footnotesize (e) }
	\end{tabular}
	\caption{Five kinodynamic motion planning problems in our benchmark \textit{Dynobench}, with a solution found by \texttt{iDb-RRT-C}.
		(a) \textit{Rotor Pole - Up obstacles 2} (b) \textit{Unicycle 2 - Narrow passage}  (c)  \textit{Car with Trailer - Double bugtrap} (d) \textit{Quadrotor v0 - Recovery obstacles 2} (e) \textit{Quadrotor v1 - Double window}.
		\vspace{-.3cm}
	}

\end{figure*}

We include a diverse range of dynamical systems and environments, featuring varying state dimensionality (from 3 to 14), the number of underactuated degrees of freedom, and controllability.
All systems use explicit Euler integration \eqref{eq:dynamics_discrete}, with \(\Delta t=\SI{0.1}{s}\) for all car-like robots and \(\Delta t=\SI{0.01}{s}\) for the flying robots and the \textit{Acrobot}.

The 8 systems are (see \cite{ortizharo2023idba} for a detailed explanation): \textit{Unicycle 1 (\(1^{\text{st}}\) order)}: 3-dimensional state space and 2-dimensional control space.
\textit{Unicycle 2 (\(2^{\text{nd}}\) order)}: 5-dimensional state space and 2-dimensional acceleration control. \textit{Car with trailer}: 4-dimensional state space and a 2-dimensional control space, \textit{Acrobot}: 4-dimensional state space and 1-dimensional control space. \textit{Quadrotor v0}: 13-dimensional state space and a 4-dimensional control space (force for each of the four motors)\footnote{We use the parameters of the Crazyflie 2.1, where the low thrust-to-weight ratio of \num{1.3} is very challenging for kinodynamic motion planning.}. \textit{Quadrotor v1}: The state space is the same as in \textit{Quadrotor v0}, but controls are now the total thrust and torques in the body frame. \textit{Planar rotor}: 6-dimensional state space and 2-dimensional control space, also with \num{1.3} thrust-to-weight ratio. \textit{Rotor pole}: 2-dimensional control space and 8-dimensional state space.


\subsection{Metrics}

Each experiment is run \num{20} times with different random seeds on a desktop computer\footnote{Intel(R) Xeon(R) W-2145 CPU @ 3.70GHz}, single-core.
We report: \begin{itemize} \item \(t [s]\): Compute time to get the first solution (median).
	\item \(c [s]\): Cost of the first solution.
	      As a cost, we use the duration of the found trajectory, in seconds (median).
\end{itemize}

If all the runs of an algorithm fail to find a solution before the timeout of \SI{60}{s}, we use a dash (`-`) in the table.
If less than \num{50}\% of the runs find a solution, we report the best value but add an asterisk (`*`) to indicate a low success rate.

\subsection{Algorithms}

We analyze two variants of the iDb-RRT family (\cshref{alg:overview}): \begin{itemize} \item \texttt{iDb-RRT-F}: using a forward Db-RRT (\cshref{alg:rrt}).
	\item \texttt{iDb-RRT-C}:
	      using a bidirectional Db-RRT inspired by RRT-Connect.
\end{itemize}

We compare our algorithms against state-of-the-art methods that use optimization, search, and sampling, and have available open-source implementations.

$\bullet$ For a sampling-based approach, we use the kinodynamic version of RRT implemented in OMPL~\cite{OMPL} (Open Motion Planning Library), which uses the propagation of random control inputs to grow the search tree.
Since sampling-based kinodynamic approaches cannot reach a goal state exactly, we use a goal region using the same value of \(\delta\) used in iDb-A* and iDb-RRT.
We denote this algorithm as \texttt{Kino-RRT}.

$\bullet$ For optimization-based planning, we choose a standard combination of a geometric motion planner and a trajectory optimizer, which we denote as  \ALGrrt.
Specifically, we use a \textit{geometric} RRT (using the implementation in OMPL) to plan using only the position and orientation of the system, without considering velocity and dynamics.
The trajectory optimizer (also based on Feasibility-driven DDP~\cite{Crocoddyl}, see \cite{ortizharo2023idba} for details) is warm-started with the geometric guess.
If trajectory optimization fails, we run RRT again from scratch and repeat.


$\bullet$ \texttt{iDb-A*} is a hybrid method that integrates search with motion primitives and trajectory optimization, but uses incremental A*-searches instead of RRT.
Notably, \texttt{iDb-A*} has been designed to combine asymptotic optimality with good anytime behavior, as it starts with a small number of motion primitives and incrementally increases the number of available motion primitives during each A*-search.
We terminate the algorithm once the first solution is found.




In all algorithms, all hyperparameters are chosen per dynamical system.

\subsection{Results -- Comparison with Baselines}

\begin{table*}
	\centering
	\scriptsize
	\caption{
		\vspace{-.1cm}
		Median initial solution time (t) and median initial cost (c) for the benchmarked systems and algorithms. 
		 \vspace{-.2cm}
	}
	\label{tab:all}


	\begin{tabular}{lcccccccccc}
\toprule
 Problem  &\multicolumn{2}{c}{iDb-RRT-F}&\multicolumn{2}{c}{iDb-RRT-C}&\multicolumn{2}{c}{Geo-RRT-TO}&\multicolumn{2}{c}{iDb-A*}&\multicolumn{2}{c}{Kino-RRT}\\
\cmidrule(lr){2-3}\cmidrule(lr){4-5}\cmidrule(lr){6-7}\cmidrule(lr){8-9}\cmidrule(lr){10-11}
	  &  t [s] & c [s] & t [s] & c [s] & t [s] & c [s] & t [s] & c [s] & t [s] & c [s]\\
\midrule
Acrobot/Swing up & 0.35 & 5.39 & { \bf 0.25} & 5.95 & 0.82 & { \bf 4.21} & {  1.49} & 5.53 & 0.32 & {  6.83}\\
Acrobot/Swing up obstacles v1 & 0.36 & 5.37 & { \bf 0.18} & { \bf 4.86} & 0.80 & 5.06 & {  1.92} & 5.80 & 0.38 & {  6.19}\\
Car with trailer/Kink & 0.23 & 53.05 & 0.24 & 60.85 & 0.59 & 34.45 & {  1.29} & { \bf 31.10} & { \bf 0.20} & {  68.50}\\
Car with trailer/Park & {  0.10} & 10.85 & { \bf 0.05} & 14.00 & 0.10 & { \bf 5.05} & {  0.11} & {  17.90} & { \bf 0.05} & 8.15\\
Planar rotor/Hole & { \bf 0.56} & 8.88 & 1.00 & 10.93 & 8.63 & 5.47 & 11.77 & { \bf 3.49} & 3.04* & 5.99*\\
Planar rotor/Bugtrap & 1.44 & 9.97 & { \bf 1.04} & 10.48 & 0.46* & 7.84* & 12.79 & { \bf 5.17} & 39.23 & 10.55\\
Rotor pole/Swing up obstacles & 1.91 & 8.20 & { \bf 1.14} & 8.38 & 10.70 & 6.09 & 2.96 & { \bf 3.98} & - & -\\
Rotor pole/Small window & 3.84 & 9.43 & { \bf 1.14} & 9.30 & 6.21* & 2.99* & 4.39 & { \bf 4.54} & - & -\\
Quadrotor v0/Recovery & 0.83 & 5.61 & { \bf 0.71} & 5.25 & 1.12 & { \bf 2.53} & 1.32 & 5.57 & - & -\\
Quadrotor v0/Recovery obstacles & 1.29 & 6.41 & 1.37 & 6.20 & { \bf 0.71} & { \bf 3.90} & 1.53 & 5.72 & - & -\\
Quadrotor v1/Obstacle & 0.87 & 6.00 & 1.36 & 7.03 & { \bf 0.25} & { \bf 2.72} & 2.53 & 4.54 & 40.68* & 4.90*\\
Quadrotor v1/Window & { \bf 0.61} & 5.22 & 0.88 & 7.99 & 9.05 & 5.53 & 1.64 & { \bf 3.71} & 9.83* & 10.08*\\
Unicycle 1 v0/Bugtrap & 0.13 & 33.05 & { \bf 0.11} & 30.45 & 0.40 & 40.35 & {  0.52} & { \bf 22.20} & 0.14 & {  70.30}\\
Unicycle 1 v2/Wall & 0.09 & 30.70 & { \bf 0.04} & 31.95 & 0.91 & 24.30 & {  0.94} & { \bf 19.60} & 0.24 & {  49.45}\\
Unicycle 2/Bugtrap & 0.16 & 59.65 & { \bf 0.09} & 56.35 & 0.61 & 43.50 & {  1.65} & { \bf 25.30} & 0.18 & {  69.25}\\
Unicycle 2/Park & 0.03 & 12.20 & { \bf 0.01} & 9.85 & {  0.12} & 6.15 & { \bf 0.01} & { \bf 5.80} & 0.05 & {  13.20}\\
\midrule
Car with trailer/Double bugtrap & 0.70 & 93.90 & { \bf 0.65} & 101.60 & 2.44* & 53.00* & 1.71 & { \bf 46.80} & 3.63 & 96.65\\
Car with trailer/Narrow passage & { \bf 0.57} & 122.55 & 0.62 & 132.05 & 0.82* & 74.50* & 8.61 & { \bf 53.90} & 2.33 & 136.25\\
Planar rotor/Recovery obstacles 2 & { \bf 0.48} & 10.75 & 0.52 & 10.95 & 6.54* & 8.94* & 20.39 & { \bf 6.04} & - & -\\
Planar rotor/Double bugtrap & 1.97 & 14.05 & { \bf 1.84} & { \bf 13.78} & - & - & - & - & 19.49* & 10.30*\\
Rotor pole/Up obstacles 2 & 4.94 & 10.68 & { \bf 2.11} & 12.15 & - & - & 14.80 & { \bf 5.00} & - & -\\
Rotor pole/Small window 2 & 3.87 & 11.59 & { \bf 1.87} & 11.91 & 0.32* & 5.71* & 11.84 & { \bf 6.18} & - & -\\
Quadrotor v0/Double bugtrap 3D & 5.80 & 11.87 & { \bf 4.43} & 13.19 & - & - & 23.42 & { \bf 6.36} & - & -\\
Quadrotor v0/Recovery obstacles 2 & { \bf 2.50} & 9.74 & 2.58 & 10.26 & 0.33* & 3.90* & 6.49 & { \bf 6.41} & - & -\\
Quadrotor v1/Recovery obstacles 2 & 2.50 & 9.71 & { \bf 2.38} & 9.68 & 0.30* & 3.72* & 59.71 & { \bf 6.23} & - & -\\
Quadrotor v1/Double Window & { \bf 2.18} & 7.79 & 2.54 & 10.86 & 0.42* & 4.62* & 25.74 & { \bf 5.09} & - & -\\
Unicycle 1 v0/Double bugtrap & 0.23 & 60.10 & 0.21 & 60.10 & {  1.70} & 64.75 & 0.92 & { \bf 30.10} & { \bf 0.17} & {  109.30}\\
Unicycle 1 v0/Narrow passage & 0.22 & 81.40 & { \bf 0.17} & 90.35 & 1.14 & 83.90 & {  1.88} & { \bf 37.30} & 0.20 & {  133.55}\\
Unicycle 2/Double bugtrap & 0.30 & 94.45 & 0.28 & 90.40 & 2.21 & 70.85 & {  2.37} & { \bf 34.60} & { \bf 0.25} & {  103.30}\\
Unicycle 2/Narrow passage & 0.31 & 122.50 & { \bf 0.24} & 118.20 & 1.11 & 84.30 & {  7.30} & { \bf 42.70} & 0.34 & {  124.95}\\
\bottomrule
\end{tabular}

	\vspace{-.2cm}
\end{table*}


%


Results are summarized in \cref{tab:all}.
Due to space constraints, we report only the median of each metric.
A graphical representation of these results using boxplots is available on our website.
In general, we observe that \texttt{iDb-A*} has lower variance than \texttt{iDb-RRT}, \texttt{Kino-RRT}, and \texttt{Geo-RRT-TO}.
\texttt{iDb-RRT-F} and  \texttt{iDb-RRT-C} solve all problems with a success rate of \num{100}\% (except two problems each, where they achieve \num{80}-\num{90}\% success rate), outperforming all baseline algorithms in terms of compute time to generate a solution (e.g., \texttt{iDb-RRT-C} is the fastest in \num{19} problems,
and \texttt{iDb-RRT-F} is the fastest in \num{6} problems).

\begin{itemize}
	\item \texttt{Kino-RRT}: it finds a first solution
	      in low-dimensional car-like systems in a competitive timeframe (but slower than \texttt{iDb-RRT-F} in \num{13} out of \num{19} cases) with a higher average cost.
	      However, in agile systems (e.g., flying robots), propagation of random control inputs is very inefficient, and \texttt{Kino-RRT} fails to find a solution in \num{11} problems out of \num{30}.

	\item \texttt{Geo-RRT-TO} often requires multiple runs of RRT to provide a suitable initial guess for trajectory optimization, and sometimes fails completely as the initial guesses never contain information about the dynamics of the system (solving only \num{18} out of \num{30} problems with a success rate above \num{50}\%).
	      If the initial guess works for the optimizer, it can be very fast (\texttt{Geo-RRT-TO} is faster than \texttt{iDb-RRT-F} in \num{7} problems).

	\item \texttt{iDb-A*}: is the strongest baseline, with success rate of \num{100}\% in all problems except \textit{Planar rotor/Double bugtrap}.
	      However, \texttt{iDb-A*} is always outperformed in the time to find the first solution by \texttt{iDb-RRT-C}.
	      The difference between \texttt{iDb-A*} and \texttt{iDb-RRT-C} increases in the new benchmark (last 14 problems), which require longer plans, with improvements up to \num{10}-\num{20}{x}.
	      On the other hand, the first solution found with \texttt{iDb-A*} has a better cost than any other algorithm in \num{23} cases.


\end{itemize}

%
%
%

\subsection{Discussion}

\paragraph{Forward vs Bidirectional Search}

Comparing our two variants, we observe that \texttt{iDb-RRT-C} is better in \num{21} out of \num{30} problems in terms of compute time.
These results agree with previous experiments in the RRT literature, where RRT-Connect is generally faster than a forward search (in robotics problems, starting a search from the start and the goal is often beneficial because these configurations are often close to obstacles and narrow passages).

\paragraph{Number of primitives and discontinuity bound}

Connecting primitives with discontinuities allows our algorithms to plan using a reduced number of primitives.
As a reference, for the system \textit{Unicycle 1 v0}, we use an initial set of 200 primitives and an initial discontinuity bound of \num{0.3}.
The discontinuity is computed with a weighted Euclidean norm (e.g., weight \num{1} for position and \num{0.5} for orientation); thus a $\delta$ of $0.3$ could represent up to \SI{30} {cm} of discontinuity in position or \SI{0.6}{rad} in orientation.
Such discontinuities are large enough that the trajectory is not directly applicable to the real robot, but it can be efficiently repaired in the nonlinear trajectory optimization step of iDb-RRT.
For the \textit{Quadcopter v0}, we use \num{5000} primitives and a discontinuity bound of \num{0.35}, and for the \textit{Rotor pole}, we use \num{8000} motions and a discontinuity bound of \num{0.45}.
The time spent to generate one motion primitive (offline) ranges from \num{10}{ms} for car-like robots to up to \num{5}{s} for flying robots (most of the time is spent attempting to solve two-point boundary value problems that do not have a solution).


\paragraph{Analysis of compute time in iDb-RRT}
In iDb-RRT, the time spent in trajectory optimization dominates the total compute time.
For instance, the compute time required to optimize one trajectory in the new benchmark problems with flying robots is between \num{1}{s} and \num{3}{s}, while in car-like robots is between \num{50}{ms} and \num{200}{ms}.
In addition, in the systems \textit{Quadrotor v0} and \textit{Quadrotor v1}, trajectory optimization may fail at the first attempt, and finding a feasible solution requires multiple iterations of iDb-RRT.
For car-like systems, we can compute trajectories of duration up to \num{50}{s} in less than \num{1}{s}. 
For flying robots, we require between \num{0.5}{s} and \num{4}{s} to generate trajectories of duration up to \num{14}{s}.
A straightforward way to speed up the trajectory optimization step is to reduce the time discretization from \num{0.01}{s} to \num{0.05}{s} (with an expected \num{5}{x} speedup).

\paragraph{RRT is easier to tune than incremental A*}


The running time of A* with motion primitives in a continuous space is highly sensitive to the number of motion primitives, i.e., the discretization level.
With too few primitives, the problem becomes unsolvable; with too many, the state space to be expanded becomes unmanageably large.
Conversely, our iDb-RRT algorithms lack an explicit notion of a branching factor.
As confirmed by our results, the RRT approach naturally adapts to efficiently solving both simple and complex problems alike, obviating the need for choosing a branching factor while also providing faster exploration.



\paragraph{Limitations and future work}

The main limitation of iDb-RRT, similar to iDb-A\*, lies in its scalability to higher-dimensional systems.
As the dimensionality increases, the number of motion primitives required to cover the state space with a small discontinuity grows exponentially.
This issue can be partially mitigated by planning with larger discontinuities.
In our benchmark, we successfully scaled to 13-DOF for the \textit{Quadrotor} and 8-DOF for \textit{Rotor pole}, thanks to leveraging translation invariance and the second-order linear velocity invariance of the dynamics.
To effectively scale to higher dimensions, we see great potential in using function approximation to learn a more informative distance metric and to combine motion primitives with deep generative models or learned policies.

\section{Conclusion}

We present iDb-RRT, a novel algorithm for kinodynamic motion planning that combines search and optimization within the framework of Rapidly-Exploring Random Trees (RRT).
Our algorithm connects motion primitives with a bounded discontinuity as the expansion step of an RRT, which is later repaired using trajectory optimization.
iDb-RRT is probabilistically complete and finds solutions faster than state-of-the-art kinodynamic motion planning across a diverse set of problems.

Comparatively, iDb-RRT and iDb-A* possess complementary strengths: the former finds solutions significantly faster, while the latter converges to optimal solutions with more compute time.
Together, they demonstrate that combining motion primitives, bounded discontinuity, and trajectory optimization, is a promising approach for both sampling-based and search-based motion planning.









\bibliographystyle{IEEEtran}
\bibliography{IEEEabrv,refs}

\end{document}